# Fast geodesic shooting for landmark matching using CUDA

Jiancong Wang

**Abstract.** Landmark matching via geodesic shooting is a prerequisite task for numerous registration based applications in biomedicine. Geodesic shooting has been developed as one solution approach and formulates the diffeomorphic registration as an optimal control problem under the Hamiltonian framework. In this framework, with landmark positions $q_0$ fixed, the problem solely depends on the initial momentum $p_0$ and evolves through time steps according to a set of constraint equations. Given an initial $p_0$, the algorithm flows $q$ and $p$ forward through time steps, calculates a loss based on point-set mismatch and kinetic energy, back-propagate through time to calculate gradient on $p_0$ and update it. In the forward and backward pass, a pair-wise kernel on landmark points $K$ and additional intermediate terms have to be calculated and marginalized, leading to $O(N^2)$ computational complexity, $N$ being the number of points to be registered. For medical image applications, $N$ maybe in the range of thousands, rendering this operation computationally expensive. In this work we propose a CUDA implementation based on shared memory reduction. Our implementation achieves nearly 2 orders magnitude speed up compared to a naive CPU-based implementation, in addition to improved numerical accuracy as well as better registration results.

## 1    Introduction

Diffeomorphic landmark matching introduced in [1] has been solved using geodesic shooting, [2] following a simplified formulation based on Hamiltonian optimal control. The geodesic formulation ensures the smoothness of the transformation and prevents potential folding of the manifold. In addition, the high-dimensional dense time varying velocity fields can be compactly represented by the initial momentum at the initial time point that lies in a low-dimensional linear space [2], [3]. When fitting a template to a group of subjects, the population's anatomical variability can subsequently be characterized using linear statistical methods, such as principal component analysis [4], principal geodesic analysis [5] and others [6]. This is critical for image-based studies of disease because it can be used to capture the characteristics of anatomical variability between disease cohort and normal controls.

Despite its favorable theoretical properties and relevance to small shape statistical analysis, landmark matching via geodesic shooting has not been widely used in large-scale population studies or clinical applications, because of its computational cost. The basic algorithm has computation complexity of $O(N^2)$ that scales quadratically with number of landmark $N$. However, graphical processing unit (GPU) has been shown to be effective in speeding up related algorithm [7-9]. In this work, pursue this promising direction to achieve speed up of the geodesic shooting approach to landmark matching by implementing the algorithm using CUDA (NVIDIA, Santa Clara, CA). An implementation with cuBLAS (NVIDIA, Santa Clara, CA) is first presented and memory footprint and access issues of this approach are discussed; these challenges motivate, a second, more efficient design based on reduction using shared memory. Preliminary data are presented demonstrating the potential of the proposed GPU-based implementations.

## 2 Method

### 2.1 Landmark matching via geodesic shooting

We adopt [2]'s notation to describe the geodesic shooting algorithm. Let $X_j$ be the set of template landmarks and $X_m$ be the corresponding set of target landmarks. Let $N$ be the number of landmarks. The algorithm finds a spatial transformation $\phi$ that optimally matches the landmarks $X_j$ to $X_m$ under a regularization constraint.

Let the positions of the landmarks be a function of time ($t \in [0, 1]$) $q(t) = \{q_1(t), \ldots, q_L(t)\}$ and the momentum of the landmarks be $p(t) = \{p_1(t), \ldots, p_L(t)\}$ ($p_i(t) \in \mathbb{R}^3$).

The initial positions of the landmarks are given $q(0) = X_m$, and the initial momenta of the landmarks $p(0) = \alpha_j^m$ are an unknown that the geodesic shooting algorithm optimizes for.

The evolution of the system is formulated in terms of the Hamiltonian:

$$H(p, q) = p, K(q) p > (1)$$

H(p, q) is the kinetic energy of the system and is constant over time when the evolution of the system is formulated in terms of the Hamiltonian:

$$\begin{cases} \dfrac{dq}{dt} = \dfrac{\partial H}{\partial p}(q, p) \\ \dfrac{dp}{dt} = \dfrac{-\partial H}{\partial q}(q, p) \end{cases} (2)$$

where $K(q)$ is a $(3N) \times (3N)$ with $N \times N$ diagonal blocks, with the $(l, n)$-th block equal to $G\sigma_{gs}(|| q_l - q_m ||) \cdot I_3$, $G\sigma_{gs}$ being a Gaussian kernel and $I_3$ being an identity matrix.

The landmark matching problem is formulated as the optimization

$$\alpha^\square = \underset{\alpha \in R^{3L}}{argmax}\, H(X^m) + \lambda \cdot \left\| q(1) - X^j \right\|_2^2 (3)$$

This minimization problem is discretized in time and solved using gradient-based optimization [2]. In our implementation, we use L-BGFS [10] from VNL numeric library [11] to perform second order update.

Given the optimal solution $\alpha^*$, and the corresponding landmark trajectories $q^*(t)$ and momenta $p^*(t)$ the landmark transformation can be interpolated over the entire spatial domain to yield a smooth velocity field

$$v(x,t) = \sum_{l=1}^{L} G_{\sigma_g}\left(\left\|q_l^\square - x\right\|_2\right) \cdot p_l^\square(t), x \in R^3 \quad (4)$$

In summary, the geodesic shooting algorithm yields an initial momentum $\alpha^*$, which can be converted to diffeomorphic transformation $\phi$ that approximately matches landmarks $X_m$ to landmarks $X_j$, subject to a regularization term that penalizes the kinetic energy of the landmark transformation.

## 2.2 CPU-based implementation

We have implemented and validated the landmark matching geodesic shooting algorithm on CPU in prior work. This implementation serves as a baseline for further optimization and benchmarking, and its pseudocode is given in Algorithm 1.

---

**Algorithm 1:** LANDMARK MATCHING BY GEODESIC SHOOTING

**Input:** Template landmark position $q_0$, Target landmark position $q_T$, Total Timestep $T$, Gaussian std $\sigma$, Trade off $\lambda$, Optimization Iteration $I$

**Output:** Initial momentum $p_0$

1 **Function** $ComputeGradient$ $(q_0, q_T, T, \sigma, \lambda, I)$
2     $d_t \leftarrow 1.0/T$
3     $p_t \leftarrow p_0$
4     $q_t \leftarrow q_0/T$
5     **for** $t \leftarrow 1$ **to** $T$ **do**
6        $Hq, Hp \leftarrow StepForward(q_t, p_t, q_T, \sigma, \lambda)$
7        $q_t \leftarrow qt + Hp * dt$
8        $p_t \leftarrow pt - Hq * dt$
9     $q_1 \leftarrow q_t$
10     $loss = \lambda * sum((q_t - q_1)^2) + velocity(p_t, q_t)^2$
11     $\alpha_t, \beta_t \leftarrow derivative(loss, q_t, p_t)$
12     **for** $t \leftarrow T$ **to** $1$ **do**
13        $d\alpha, d\beta \leftarrow StepBackward(\alpha_t, \beta_t, \sigma, \lambda)$
14        $\beta_t \leftarrow \beta_t + d\beta * dt$
15        $\alpha_t \leftarrow \alpha_t - d\alpha * dt$
16     $dp_0 \leftarrow \alpha_t$
17     **return** $dp_0$
18 $p_0 \leftarrow (q_T - q_0)/T$
19 $LBGFS(p_0, ComputeGradient, I)$
20 **return** $p_0$

---

In summary, the algorithm performs the following steps.

(a) Initialize $p_0$.

(b) Flow $q$ and $p$ forward through time by equation (2) in section 2.1.

(c) Calculates a loss based on landmark mismatch and kinetic energy.

(d) Back-propagate through time to calculate gradient on $p_0$.

(e) Run (a-d) multiple times to perform line search and L-BGFS update.

(f) Repeat (e) until convergence or until maximum iteration set by the user.

In the pseudocode, the *StepForward* function carries out the flow forward operation (b) while the *StepBackward* function carries out the back-propagate operation (d).

We denote the two derivative terms in equation (2) as $H_p$ and $H_q$. The *StepForward* function calculates the Hamiltonian and the two derivatives $H_p$ and $H_q$, which requires calculation of $K$, individual $H_q$, $H_p$ terms based on $K$, and the marginalization across them. $K$ is matrix of size $N^2$ and exact calculation is of time complexity $O(N^2)$.

Algorithm2 contains the pseudocode for CPU-based implementation of the *StepForward* function. As the pseudocode shows, this implementation uses a double for loop to accumulate individual terms into *Hq* and *Hp*. *N*, the number of landmarks, may typically be in the thousands, rendering this operation computationally expensive. To compound the problem, the *StepForward* function runs in each time step/line-search procedure in L-BGFS/iteration, which adds up to a constant factor of hundreds of thousands. The *StepBackward* function has similar computation complexity. Optimizing these functions is therefore key to accelerating the overall algorithm.

---

**Algorithm 2:** STEPFORWARD

    **Input:** Template landmark position at time t $q_t$, Target landmark
              position $q_T$, Momentum at time t $p_t$, Gaussian std $\sigma$, Trade off $\lambda$,
              number of landmark $N$

    **Output:** Geodesic update to p and q $H_q, H_p$

1  **Function** *StepForward* $(q_0, q_T, \sigma, \lambda, I)$

2     |  $H_q \leftarrow 0.0$

3     |  $H_p \leftarrow 0.0$

4     |  **for** $i \leftarrow 1$ ***to*** $N$ **do**

5     |    |  **for** $j \leftarrow 1$ ***to*** $N$ **do**

6     |    |    |  $K = exp((q_t[i] - q_t[j])^2 / \sigma)$

7     |    |    |  $Hq[i] \leftarrow Hq[i] + precomputeHqTerms(K, q_t, p_t, q_T, \sigma, \lambda)$

8     |    |    |  $Hp[i] \leftarrow Hp[i] + precomputeHpTerms(K, q_t, p_t, q_T, \sigma, \lambda)$

9   **return** $H_q, H_p$

---

There are two immediate problems with the CPU-based implementation. First this implementation directly accumulates into $H_p$ and $H_q$. They are large arrays residing in RAM, which are expensive to access. Instead the accumulation could be done on a local variable and be written to host memory only once at the end of the inner loop, since a local variable access will most likely be compiled into a register access.

Second, using a for loop to sequentially add thousands of terms to one variable leads to low numerical accuracy, since in the late iterations, the scale of the sum will be much larger than individual terms. This is particularly problematic when single precision is used, as shown in the results section.

## 2.3 CUDA programming model and memory hierarchy

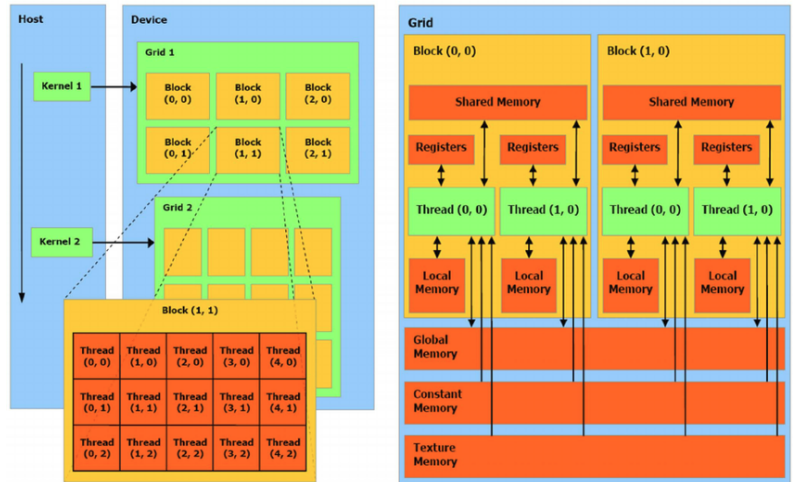

Fig. 1

Schematization of CUDA architecture. Figure taken from Nvidia CUDA C Programming Guide v.5.0, 2012 [12]. Left: Thread arrangement in CUDA. Right: CUDA memory hierarchy.

CUDA is an extension to C/C++ for programming on the Nvidia GPU [12]. Since GPU is naturally multi-core, CUDA programs, or kernels, are designed to be run with multiple threads. Threads in CUDA are organized into thread blocks, blocks in turn are organized in grids (Figure 1, left). Blocks and threads can be up to 3 dimensions. Each thread/block is given index numbers and the kernel runs on all of them, with thread behavior varying mainly by the index number to collectively accomplish tasks.

In CUDA programming model, memory is roughly organized into 3 types: host memory, device global memory, and device local memory (Figure 1, right). Host refers to the CPU; device refers to GPU; global refers to memory that all GPU threads can access and local refers to memory only a thread or thread block can access. Speed is roughly in the following order: CPU memory $\ll$ device global memory $\ll$ device local memory, where $\ll$ indicate 20-100 times difference. Size is roughly in the opposite order. Optimization in CUDA programming aims to reduce an algorithm's global memory footprint and usage in favor of shared memory and register access.

## 2.4 Implementation of *StepForward* and *StepBackward* using cuBLAS

cuBLAS is a fast BLAS library provided by CUDA. It is significantly faster than normal CPU BLAS [13], making it an obvious implementation option for the marginalization. The idea is to pre-calculate all the individual $H_q$ and $H_p$ terms, put them into a square matrix, and then perform the marginalization by multiplying the matrix with a $1 \rightarrow$ vector.

---

**Algorithm 3:** STEPFORWARD WITH CUBLAS

---

**Input:** Template landmark position at time t $q_t$, Target landmark
position $q_T$, Momentum at time t $p_t$, Gaussian std $\sigma$, Trade off $\lambda$,
number of landmark $N$

**Output:** Geodesic update to p and q $H_q, H_p$

1 **Function** *StepForward* $(q_0, q_T, \sigma, \lambda, I)$

2      $H_q \leftarrow 0.0$

3      $H_p \leftarrow 0.0$

4      $one\_vector \leftarrow 1.0$

5      $MemAlloc(H_q\_precompute, k * k)$

6      $MemAlloc(H_p\_precompute, k * k)$

7      $H_q\_precompute, H_p\_precompute \leftarrow$
     $H_q\_H_p\_pre\_computeKernel(q_t, p_t, q_T, \sigma, \lambda)$

8      $cublasSgemv(H_q\_precomputed, one\_vector, H_q)$

9      $cublasSgemv(H_p\_precomputed, one\_vector, H_p)$

10      $cudaFree(H_q\_precompute)$

11      $cudaFree(H_p\_precompute)$

12 **return** $H_q, H_p$

---

Numerically this approach produces almost identical results to the CPU-based implementation, but achieves roughly 25-30 times speed up. Nevertheless, two major issues afflict this approach.

The first and most serious problem is the method requires $N^2$ storage space for the intermediate $H_q$ and $H_p$ terms. Landmark set, containing, for example, 1500-2500 points, translates to a square matrix of 2-5 million float variables, or up to 2 gigabytes of memory.

Another problem with this approach is slowed speed due to unnecessary global write/read operations. Each intermediate term requires exactly one sequenced global write to the precompute matrix and later the cuBLAS reads them once again. This corresponds to $N^2$ global read/write operations.

### 2.5 Implementation of *StepForward* and *StepBackward* using shared memory reduction

First, in both *StepForward* and *StepBackward*, the quantity of concern is the sum of pair-wise terms, not the individual terms. Second, pairwise term in $K$ can be calculated on the fly by fetching directly from $p$ and $q$. These conditions enable our use of the shared memory-based reduction strategy to achieve greater speed up.

As noted before, in CUDA programming models, threads are arranged into blocks and each block in Nvidia's GPU contains a 48KB shared memory that all threads in same thread block can jointly access at low cost. This local shared memory is the key to speed up: it allows us to calculate the sum of terms locally in the shared memory and write only the sum instead of individual terms to the global memory, thus com-pletely avoiding the memory usage and read/write operation of the individual cross terms.

The shared memory for reduction is introduced in detail in the Nvidia SDK literature [14, 15]. It is a classic binary reduction scheme in CUDA, where threads pairs up to reduce the sum in shared memory (thread 0-127 sums threads 128-256 in first round, threads 0-63 sums threads 64-127 in second round, etc, repeated until only threads 0 remains). The actual implementation also involves thread shuffling to speed up last reductions.

A minor point is that in CUDA kernels for loop leads to severe thread diverging [16] and it is common practice to unloop when possible. Our implementation follows this practice and all 2D/3D functions are thus separately coded. No templating is used.

---

**Algorithm 4:** STEPFORWARD WITH SHARED MEMORY REDUCTION

**Input:** Template landmark position at time t $q_t$, Target landmark position $q_T$, Momentum at time t $p_t$, Gaussian std $\sigma$, Trade off $\lambda$, number of landmark $N$

**Output:** Geodesic update to p and q $H_q, H_p$

1 **Function** $StepForward\ (q_0, q_T, \sigma, \lambda, I)$

2     **Function** $Device\ sharedMemReductionKernel\ (H_q, H_p, q_t, p_t, \sigma, \lambda, N)$

3         $SharedMemAlloc(localSum, 256*2)$

4         $K = exp((q_t[threadIdx.x] - q_t[threadIdx.y])^2/\sigma)$

5         $localSum[threadIndex.y] = pre\_computeHqHp(q[threadIndex.y], p[threadIndex.y], \sigma, \lambda, K)$

6         $syncThreads()$

7         **for** $i \leftarrow 256, 128, 64, ...1$ **do**

8             **if** $threadIdx.y < i/2$ **then**

9                 $localSum[threadIdx.y] \leftarrow localSum[threadIdx.y] + localSum[threadIdx.y + i/2]$

10             $syncThreads()$

11         **if** $threadIdx.y == 0$ **then**

12             $atomicAdd(Hq[threadIdx.x], localSum[0].x)$

13             $atomicAdd(Hp[threadIdx.x], localSum[0].y)$

14     $H_q \leftarrow 0.0$

15     $H_p \leftarrow 0.0$

16     $sharedMemReductionKernel(H_q, H_p, q_t, p_t, \sigma, \lambda, N)$

17 **return** $H_q, H_p$

## 3    Evaluation

### 3.1    Experimental design

The developed landmark matching algorithms are tested on data extracted from human medial temporal lobe. This dataset consists of 34 T1-weighted MRI scans with corresponding manual segmentation. A template is derived from the manual segmentations [17] by applying a minimum spanning tree based template building algorithm [18]. 1847 landmarks are sampled from the template mesh using Poisson Disk sampling algorithm [19]. Then, the template landmarks are warped back to the space of each manual segmentation using the deformation field generated from the template-building step. Prior to deformable matching, the target landmarks are rigidly aligned with the Procrustes algorithm [20]. A template landmark set is created by averaging the aligned target landmarks [4]. We registered the template landmarks to each of the subject (target) using both CPU-based and GPU-based version. The registration result is evaluated using average and max pointwise distances.

$$AverageDist\left(q_1, q_T\right) = \frac{1}{N} \sum_{i=1}^{N} \left\| q_1(i) - q_T(i) \right\|_2$$

$$MaxDist\left(q_1, q_T\right) = \max_{i \leq i \leq N} \left\| q_1(i) - q_T(i) \right\|_2$$

The algorithm parameters used in the evaluation are as follows: Gaussian $\sigma = 1.5$; time steps $T = 40$; maximum iteration $I = 400$; weighting parameter $\lambda = 500000$. Both CPU-based and GPU-based versions were run using single precision (float32). The CPU-based version was implemented on a cluster node (E5-2643 v3 CPU, 16GB RAM, Centos 6), and the GPU-based version on a desktop machine (I7 4790K CPU, GTX 980TI GPU, 32GB RAM, Ubuntu 16.04). The CPU-based version was compiled with gcc 4.4.7 and GPU version with CUDA 8 GA2 on gcc 5.4.

### 3.2    Results

A short video is uploaded to illustrate the template landmark-set evolving through 40 time steps and into alignment with a target landmark set (registration.avi). The difference in registration accuracy between CPU-based and GPU-based implementation is due to the numerical accumulation error in CPU-based version.

**Table 1.** Comparison of performance between CPU-based and GPU-based implementations

| Implementation | Average landmark distance (mm) | Max landmark distance (mm) | Average run time |
|---|---|---|---|
| Before registration | 1.7439 | 5.7804 | - |
| CPU | 0.9001 | 3.0685 | 35±5min (65x) |
| GPU | **0.0890** | **0.4690** | **32±3sec (1x)** |

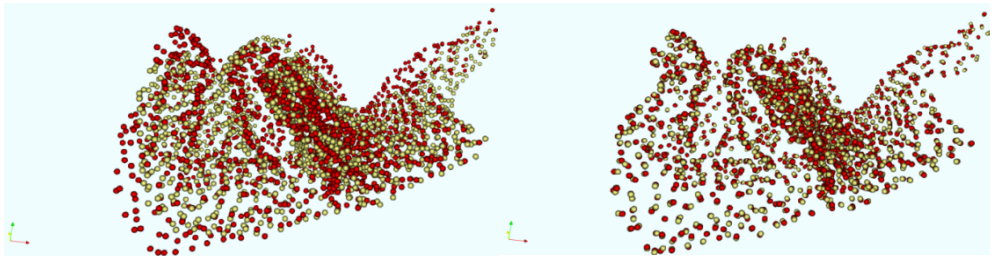

**Fig. 2** Example registration result. Left. The template landmark-set (yellow) and target landmark-set (red) before registration. Right. After GPU-based registration.

## 4    Conclusion

A CUDA implementation of the geodesic shooting approach for landmark based matching is presented in this paper. The registration problem is formulated within a Hamiltonian optimal control framework across multiple time steps. The result is an iterative update formula on landmark positions and momentums. The kernel terms in the require quadratic on point number, thus presenting a major computational challenge.

To address this and reduce memory footprint and access, the classic binary reduction scheme carried out in shared memory is used to implement the algorithm in CUDA. The result is a significant speed up over the CPU-based version or a naively implemented GPU-based version.

Specifically, compared to a naive CPU-based implementation, the shared memory reduction version is able to register a landmark set of size ~1800 on a GTX 980Ti desktop GPU, 60-70 times faster (~30 seconds), and produce more numerically accurate results as a consequence of the binary reduction scheme used. The implementation is tested on a human cortex dataset. The performance is shown to be promising. Further improvements are possible by reformulating the computation of the pair-wise kernel terms, and these are the basis for future work.